\documentclass[letterpaper]{article} 
\usepackage{aaai2026}  
\usepackage{times}  
\usepackage{helvet}  
\usepackage{courier}  
\usepackage[hyphens]{url}  
\usepackage{graphicx} 
\usepackage{natbib}  
\usepackage{caption} 
\usepackage{algorithm}
\usepackage{algorithmic}

\usepackage{newfloat}
\usepackage{listings}
\DeclareCaptionStyle{ruled}{labelfont=normalfont,labelsep=colon,strut=off} 
\floatstyle{ruled}
\newfloat{listing}{tb}{lst}{}
\floatname{listing}{Listing}
\usepackage{inputenc}
\usepackage[listings,breakable,skins]{tcolorbox}
\newtcblisting{prompttemplate}[1]{%
  title={#1},
  breakable,
  boxrule=0.5mm,
  listing only,
  listing options={
    basicstyle=\ttfamily,
    breaklines=true,
    columns=fullflexible,
    keepspaces=true,
  }
}
\newtcolorbox{metadata}[1]{%
  title={#1},                 
  breakable,                
  colframe=gray!50!black, 
  colback=white,          
  coltitle=white,         
  boxrule=0.5mm,             
  before upper={
    \ttfamily                
    \setlength{\parindent}{0pt} 
    \setlength{\parskip}{4pt plus 1pt} 
  }
}

\lstdefinestyle{codestyle}{
    language=Python,                   
    basicstyle=\ttfamily\footnotesize, 
    breaklines=true,                   
    breakatwhitespace=false,           
    columns=fullflexible,              
    keepspaces=true,                   
    tabsize=4,                         
    captionpos=b,                      
    showstringspaces=false,            
    keywordstyle=\color{blue},         
    commentstyle=\color{green!50!black}, 
    stringstyle=\color{red!60!black},  
    backgroundcolor=\color{white},     
    numbers=left,                      
    numberstyle=\tiny\color{gray},     
    xleftmargin=2em,                   
    framexleftmargin=1.5em,           
}

\tcbuselibrary{listings, breakable}
\newtcblisting{pythoncode}[1]{%
    title={#1},
    listing only,
    breakable,                        
    colframe=gray!50!black, 
  colback=white,          
  coltitle=white,         
    listing options={style=codestyle} 
}

\newtcblisting{prompttemplateModernGray}[1]{%
  title={#1},
  breakable,
  boxrule=0.5mm,
  colframe=gray!50!black, 
  colback=gray!10!white, 
  coltitle=white,         
  fonttitle=\bfseries,     
  listing only,
  listing options={
    basicstyle=\ttfamily\small, 
    breaklines=true,
    columns=fullflexible,
    keepspaces=true,
  }
}
\newtcblisting{prompttemplateWhite}[1]{%
  title={#1},
  breakable,
  boxrule=0.5mm,
  colframe=gray!50!black, 
  colback=white,          
  coltitle=white,         
  fonttitle=\scshape,
  listing only,
  listing options={
    basicstyle=\ttfamily\footnotesize,
    breaklines=true,
    columns=fullflexible,
    keepspaces=true,
  }
}

\usepackage{booktabs} 
\usepackage{mdframed}
\usepackage{amsmath}
\usepackage{amssymb}
\usepackage{comment}

\newcommand{\m}[1]{{\bf{#1}}}

\newcommand{\Rset}{\mathbb{R}}

\newcommand{\C}[1]{{\cal {#1}}}

\title{Toward a Trustworthy Optimization Modeling Agent \\ via Verifiable Synthetic Data Generation}
\author {
    Vinicius Lima,
    Dzung T. Phan,
    Jayant Kalagnanam,
    Dhaval Patel,
    Nianjun Zhou
}
\affiliations {
    \textsuperscript{\rm 1}IBM Research AI, Yorktown Heights, NY\\
    viniciuslima@ibm.com, phandu@us.ibm.com, jayant@us.ibm.com, pateldha@us.ibm.com, jzhou@us.ibm.com
}

\usepackage{bibentry}

\begin{document}

\maketitle

\begin{abstract}
We present a framework for training trustworthy large language model (LLM) agents for optimization modeling via a verifiable synthetic data generation pipeline. Focusing on linear and mixed-integer linear programming, our approach begins with structured symbolic representations and systematically produces natural language descriptions, mathematical formulations, and solver-executable code. By programmatically constructing each instance with known optimal solutions, the pipeline ensures full verifiability and enables automatic filtering of low-quality demonstrations generated by teacher models. Each dataset instance includes a structured representation of the optimization problem, a corresponding natural language description, the verified optimal solution, and step-by-step demonstrations — generated by a teacher model — that show how to model and solve the problem across multiple optimization modeling languages. This enables supervised fine-tuning of open-source LLMs specifically tailored to optimization tasks. To operationalize this pipeline, we introduce \texttt{OptiTrust}, a modular LLM agent that performs multi-stage translation from natural language to solver-ready code, leveraging stepwise demonstrations, multi-language inference, and majority-vote cross-validation. Our agent achieves state-of-the-art performance on standard benchmarks. Out of 7 datasets, it achieves the highest accuracy on six and outperforms the next-best algorithm by at least 8\% on three of them. Our approach provides a scalable, verifiable, and principled path toward building reliable LLM agents for real-world optimization applications. The code is available as supplementary material.
\end{abstract}


\section{Introduction}
\label{sec:intro}

Optimization serves as a foundational tool in science and engineering, underpinning a wide range of decision-making applications such as logistics, supply chain management, finance, healthcare, energy systems, and infrastructure planning. Despite its ubiquity and impact, the process of translating real-world requirements into robust, executable optimization models is often labor-intensive and demands rare expert knowledge. This modeling bottleneck limits the accessibility and scalability of optimization in practice.

Recent advances in large language models (LLMs) offer an exciting opportunity to automate the end-to-end pipeline from natural language descriptions to solver-ready code. An effective natural language-to-optimization (NL2Opt) agent has the potential to democratize optimization modeling, empowering non-experts to harness advanced mathematical tools. However, current approaches in this area face major obstacles. Many LLM-based methods produce code or models that are difficult to verify, lack transparency and reproducibility, and do not generalize well to new problem structures or domains. The scarcity of high-quality, structured datasets further impedes progress, as does the challenge of multi-stage reasoning required for accurate translation.

Building reliable optimization modeling agents presents several intertwined challenges. First, natural language problem statements are frequently ambiguous or under-specified, which means that agents must often resolve vagueness or infer missing information to produce well-posed mathematical models. Second, the translation process itself is inherently multi-stage and complex: it requires the agent to accurately identify relevant entities, assemble symbolic mathematical formulations, and ultimately generate solver-executable code that preserves the intent of the original description. Third, verifiability and fidelity remain central concerns—generated code must not only be syntactically correct but also mathematically valid and provably optimal for the given problem. Finally, the lack of large-scale, high-quality datasets for NL2Opt modeling limits the effective supervised fine-tuning of language models, since existing resources are often small, domain-specific, or lack the necessary annotations to support robust learning and evaluation. Together, these factors underscore the need for principled frameworks that enable both verifiable modeling and scalable, trustworthy LLM training for optimization tasks.

To address these barriers, we propose a novel, verifiable synthetic data generation (SDG) pipeline for training a trustworthy optimization modeling agent. Our approach begins with structured symbolic representations of optimization problems and programmatically generates aligned natural language descriptions, mathematical formulations, and solver-executable code, each instance paired with a verified optimal solution. This not only ensures data quality and full verifiability, but also enables automatic filtering of low-quality demonstrations, thereby fostering reproducible and reliable agent behavior.

\textbf{Contributions.}
Our main contributions are as follows:
\begin{itemize}
\item We design and implement a modular LLM agent, \mbox{OptiTrust}, that performs multi-stage translation from natural language descriptions to solver-ready code. \mbox{OptiTrust} generates step-by-step demonstrations for optimization modeling, leveraging multi-language inference and majority vote to improve robustness and accuracy.
\item We introduce a scalable SDG pipeline for linear and mixed-integer linear programming that creates verifiable multi-modal datasets—bridging symbolic, linguistic, and code representations—enabling fine-tuning of LLMs in five modeling languages.
\item We leverage OptiTrust to systematically identify additional inaccurate instances within 5 of the 7 existing optimization modeling datasets, particularly those containing incorrect ground-truth optimal values. By updating these values with verified solutions, we improve the reliability and quality of benchmark datasets for the community.
\item We demonstrate that our model, trained using the synthetic data pipeline, achieves state-of-the-art performance across multiple public benchmarks, outperforming all prompting and fine-tuning baselines on 6 out of 7 datasets and exceeding the next-best method by at least 8\% on three of them.
\end{itemize}

In summary, our framework closes a critical gap in the development of optimization modeling agents by combining scalable synthetic data generation with rigorous verification. This work lays the groundwork for trustworthy, reproducible, and accessible optimization modeling using LLMs.

\subsection{Related Work}
\label{sec:lit_review}

Recent work on (large) language models for optimization modeling can be broadly classified into two main directions: prompting based techniques that rely primarily on frontier models such as GPT-4 \cite{li_large_2023, xiao_chain--experts_2023, zhang-etal-2024-solving, ahmaditeshnizi_optimus_2024, astorga2025autoformulation}, and fine-tuning of open-source models using domain-specific optimization modeling datasets \cite{tang_orlm_2024, jiang2025llmopt, yang2025optibench, lu2025optmath} --- for a more comprehensive overview of natural language for optimization modeling, we refer the reader to the recent survey paper \cite{Xiao_survey_2025}.

Within the first research direction, a framework to design LLM agents to enable what-if analysis for and provide insights about existing supply-chain optimization models given natural language inputs is proposed in \cite{li_large_2023}. Chain-of-Experts \cite{xiao_chain--experts_2023} and OptiMUS \cite{ahmaditeshnizi_optimus_2024} both propose multi-agent frameworks to solve optimization problems from scratch using LLMs: in Chain-of-Experts \cite{xiao_chain--experts_2023}, each agent is assigned a specific role in the optimization pipeline (formulation, implementation and debugging, for example), while an LLM-powered orchestrator oversees the workflow used to solve the optimization problem; in OptiMUS \cite{ahmaditeshnizi_optimus_2024}, a similar framework with dedicated agents and prompts is used, and the paper further introduces a connection graph to allow independent formulation and implementation of objectives and constraints. More recently, an inference framework combining large language models and Monte Carlo tree search is proposed in \cite{astorga2025autoformulation}.

The reliance of prompt based techniques on proprietary frontier models, combined with the limited availability of datasets containing detailed problem descriptions, mathematical formulations, and solver-ready code, motivates research into synthetic data generation and training of open-source models specialized in optimization modeling. This line of work aims to both close the gap between open-source and proprietary models, and to enable smaller, more computationally efficient LLMs specialized in optimization modeling \cite{tang_orlm_2024, jiang2025llmopt, yang2025optibench, lu2025optmath}. The synthetic data generation pipelines proposed in \cite{tang_orlm_2024} and \cite{jiang2025llmopt} primarily rely on augmenting or modifying an existing pool of seed problems in natural language to generate new problem descriptions. This approach limits their ability to generate new classes of optimization problems that are not already represented among the seed problems. Furthermore, it restricts the scalability of these pipelines when handling longer descriptions or more complex problem instances. 
ReSocratic \cite{yang2025optibench} goes one step further by using a pool of formatted demonstrations as seeds for new optimization samples created by LLMs; however, there is no guarantee that the generated formulation accurately reflects the generated problem description. Concurrent to our work, recent efforts have explored similar directions on generating problem descriptions from mathematical formulations \cite{lu2025optmath}, showing promising results on leveraging structured representations of optimization models to generate optimization modeling datasets. However, existing approaches lack a mechanism for verifying the correctness of components within the pipeline. Our work is dedicated to developing a method for generating verifiable training data.  


\section{Multi-Agent Architecture for OptiTrust}
\label{sec:agent}

\begin{figure*}[ht]
\centering
\includegraphics[width=0.85\linewidth]{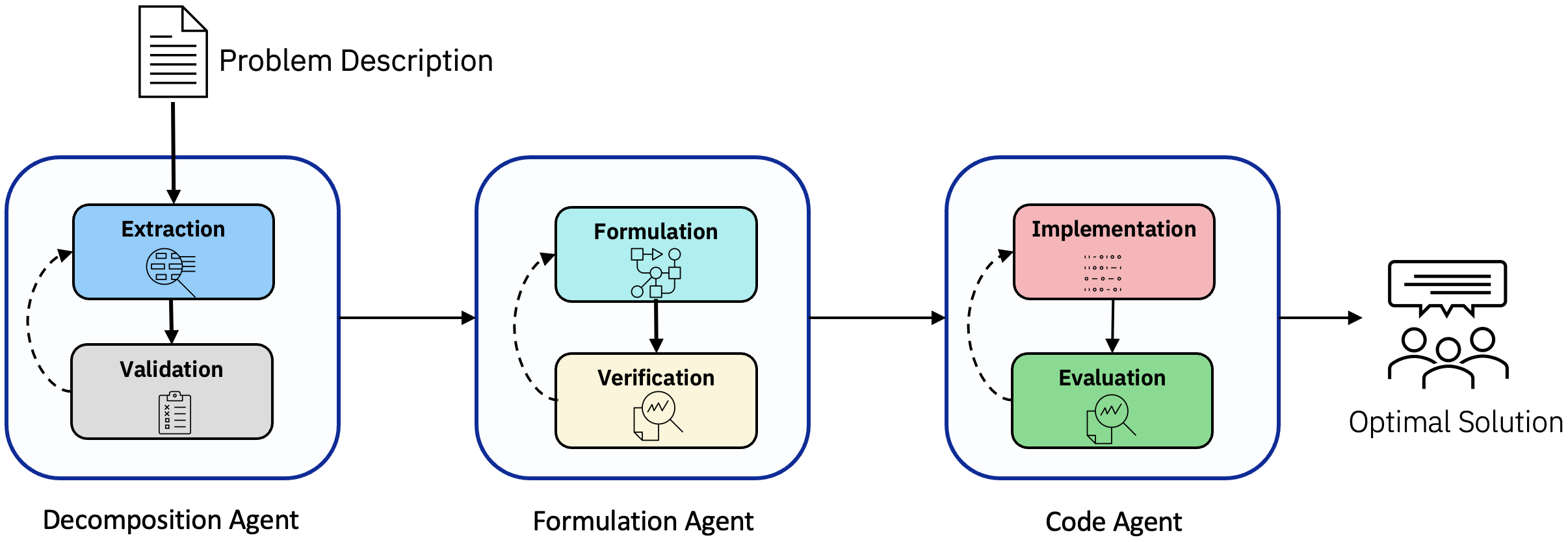}
\caption{Agentic workflow of OptiTrust from natural language to executable code.}
\label{fig:agentic-workflow}
\end{figure*}

Our proposed OptiTrust agent employs a structured, modular architecture comprising three coordinated sub-agents, each dedicated to a distinct stage in translating a natural language descriptions into solver-ready code, as shown in Figure \ref{fig:agentic-workflow}. This modular decomposition mirrors the workflow of a human optimization expert, enabling a clear division of responsibilities, improved interpretability, and effective error diagnosis. We adopt a commonly used design pattern for LLM-based agents that convert natural language descriptions into executable code for optimization solvers \cite{xiao_chain--experts_2023,ahmaditeshnizi_optimus_2024,jiang2025llmopt}, and equip each sub-agent with dedicated prompts, self-reflection, and step-by-step in-context learning demonstrations. Those step-by-step demonstrations are used not only to provide the model with explicit examples of variable definitions, problem formulations, and code implementation, but also to guide the generation of multi-task reasoning traces for supervised fine-tuning, as discussed in more details in the following section.

The \textbf{decomposition agent} initiates the workflow by parsing the natural language description provided by the user. Its core function is to identify and extract key optimization components, such as decision variables (including the domain of those variables), objectives, and constraints (including implicit ones such as non-negativity), and to summarize them using natural language. That summarized list of components, alongside the original description of the problem, is then passed to the \textbf{formulation agent}, tasked with constructing a clear, formal mathematical formulation --- typically presented in LaTeX --- that precisely captures the optimization problem. 
The workflow culminates with the \textbf{code agent}, which translates the mathematical formulation into executable optimization code (e.g., Pyomo and DOcplex) suitable for solvers such as CPLEX \cite{cplex2009v12}, Gurobi \cite{gurobi} or SCIP \cite{bolusani2024scipoptimizationsuite90}. 
An integral feature of the code agent is its built-in validation mechanism, which executes the generated code using external optimization solvers to verify correctness. If execution errors or infeasible solutions occur, detailed error messages are communicated back to the agent, triggering iterative code refinement until a viable solution is produced or a maximum number of debugging calls is reached. 
Moreover, instead of relying solely on the formulation generated in the previous step, as is usually done in the literature, the coding agent also has access to the original problem description and the detailed list of components previously extracted. That is primarily designed to allow the coding agent to retrieve information directly from the problem description in case any of the previous steps fails. 

As optimization modeling datasets are relatively scarce and often imbalanced across problem classes, and optimization modeling libraries differ in adoption as well as availability and quality of documentation, language models might be biased toward certain optimization modeling languages. 
However, existing agents are limited to working with a single modeling language. To mitigate this data imbalance and performance mismatch issue, we adopt a consistency mechanism inspired by recent works on chain-of-thought reasoning \cite{wei2022chain, wang2023selfconsistency, chen2024universal}.
In particular, we prompt the coding agent to model the problem using five common optimization modeling languages (Pyomo, Gurobipy, DOCplex, CVXPY, and PySCIPOpt) and employ a majority voting mechanism based on the solutions found by each solver to select the most consistent and reliable implementation. We find that incorporating this consistency check into the optimization workflow significantly improves model performance across both baseline and fine-tuned models, as detailed in the numerical experiments. 

The key novelty of our agent is that it incrementally evolves the components of the modeling trajectory, so if a mistake occurs at an intermediate stage, there remains an opportunity to recover the correct information from the original problem description. We provide the problem description to each sub-agent and require them to supply explicit reasoning steps. These reasoning steps, combined with majority voting based on solver code across 5 modeling languages, help promote generalization. The additional input and output components for each sub-agent are incorporated into our synthetic data, which is then used to effectively fine-tune LLMs.


\section{Verifiable Synthetic Data Generation Pipeline}

\begin{figure*}[ht]
\centering
\includegraphics[width=0.80\linewidth]{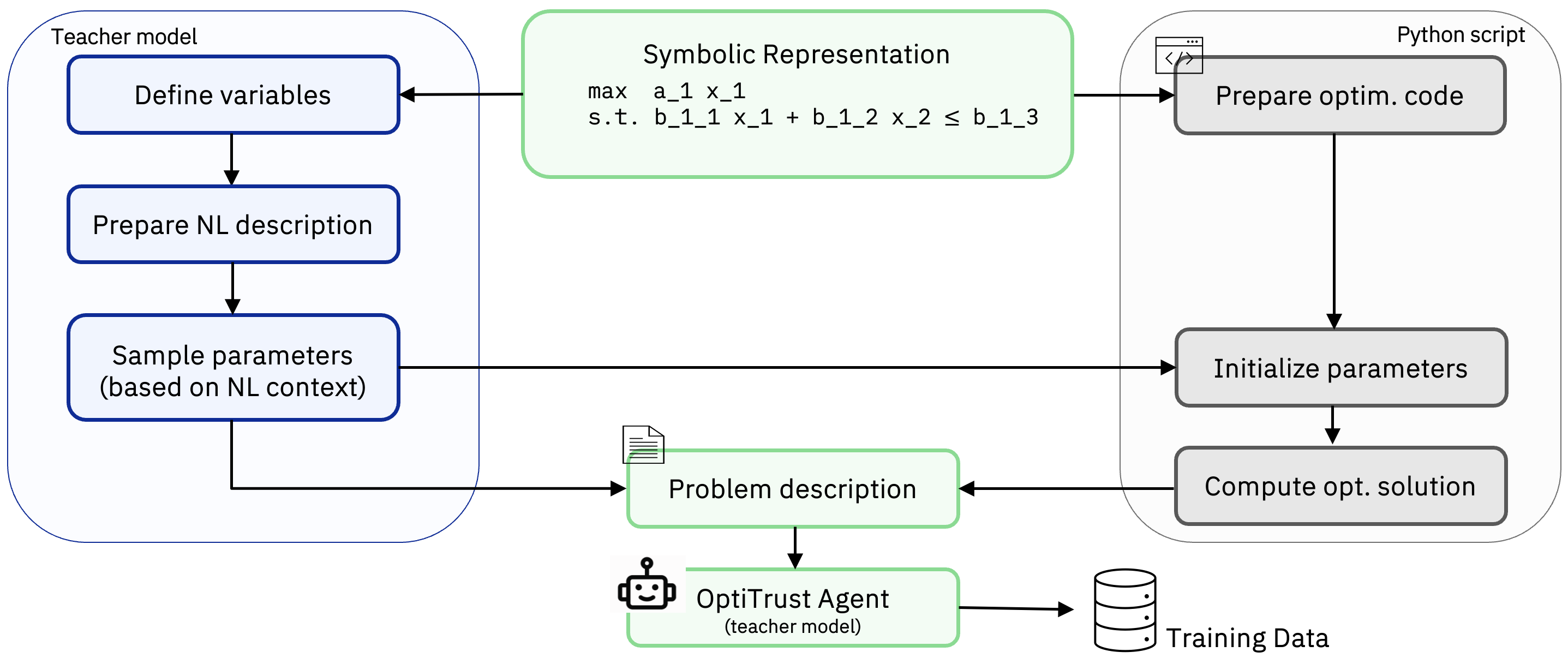}
\caption{Data generation pipeline.}
\label{fig:sdg}
\end{figure*}

Although prompt-based techniques provide flexible workflows to elicit optimization models from natural language descriptions using pre-trained LLMs, their performance ultimately depends upon the ability of pre-trained LLMs to model optimization problems and implement solver-ready code. To help bridge the performance gap between proprietary and open-source LLMs on optimization modeling tasks, on the one hand, and to enable the development of small, portable LLMs for optimization modeling, on the other, we propose a verifiable synthetic data generation pipeline to create synthetic descriptions of optimization problems. The synthetic description is then processed by the optimization workflow discussed in the previous section, using a teacher LLM model to prepare step-by-step demonstrations showing how to model optimization problems. During the data generation process, we capture reasoning traces for each step in the optimization workflow, including the implementation of optimization models in five modeling languages --- Pyomo, Gurobipy, DOcplex, CVXPY, and PySCIPOpt --- to promote robustness. The overall framework 
is illustrated in Figure \ref{fig:sdg}.


\subsection{Representing Mixed-Integer Linear Programs}

To enable automatic validation of training samples, our pipeline starts with a symbolic representation of mixed-integer linear programs (MILP). 
In a standard  MILP formulation, we model the decision-making process using a collection of decision variables, an objective function, and a set of linear constraints. The decision variables may include both continuous and integer-valued variables, representing choices such as the number of units to produce, for example, or binary on/off decisions. The objective function quantifies the underlying goal of the optimization problem -- such as minimizing cost or maximizing profit -- while a set of linear constraints captures requirements or limitations such as resource capacities, demand satisfaction, or logical relationships between decision variables. Accordingly, one can abstract a typical MILP as 
\begin{equation*}
\label{milp_model}
\begin{array}{rl l}
\C{P}: \; & \texttt{optimize}_{\m{x}\in \Rset^n} & \displaystyle\sum_{j=1}^n c_j x_j \\[2pt]
            & \text{subject to}                    & \displaystyle\sum_{j=1}^n a_{ij} x_j \; \circ \; b_i,\quad i = 1,\ldots,m \\[2pt]
            &                                      & l_j \leq x_j \leq u_j, \quad j = 1, \ldots, n \\[2pt]
            &                                      & x_j \in \mathbb{Z}, \quad j \in \mathcal{I},
\end{array}
\end{equation*}
where $\texttt{optimize} \in \{\texttt{minimize},\texttt{maximize}\}, \circ \in \{\leq, =, \geq\}, a_{ij} \in \Rset, b_i \in \Rset, l_j \in \mathbb{R} \cup \{-\infty\}, u_j \in \mathbb{R} \cup \{+\infty\}$, and $\mathcal{I}$ is the set of integer variables. 
Based on that typical representation, the SDG pipeline begins by sampling the number of decision variables $n$, the number of linear constraints $m$, and the type of optimization (i.e., $\texttt{minimize}$ or $\texttt{maximize}$), such that $\mathcal{P} = (n, m, \texttt{optimize})$. At this stage, we also sample upper or lower bounds for decision variables, if any, and the sparsity of the objective and constraint coefficients. Given the set of (symbolic) coefficients and the list of decision variables, we rely on an automated Python script to convert the symbolic representation to a parametrized Pyomo or Gurobipy template. The template defines the structure of the optimization model, but leaves coefficients and domains of variables still uninstantiated. 

\subsection{Generating Natural Language Descriptions}
While sampling the number of decision variables and constraints, as well as the sparsity of the model parameters, improves structural diversity, it does not promote semantic or linguistic diversity. To address this, we uniformly sample the problem domain from a variety of application domains (e.g., manufacturing, education, energy). We then prompt the teacher model to generate structured, concise descriptions of the decision variables, including their domain (either $\mathbb{Z}$ or $\mathbb{R}$) and implicit ranges, if any (e.g., non-negative). 

\begin{figure}[t]
\centering
\includegraphics[width=1.01\linewidth]{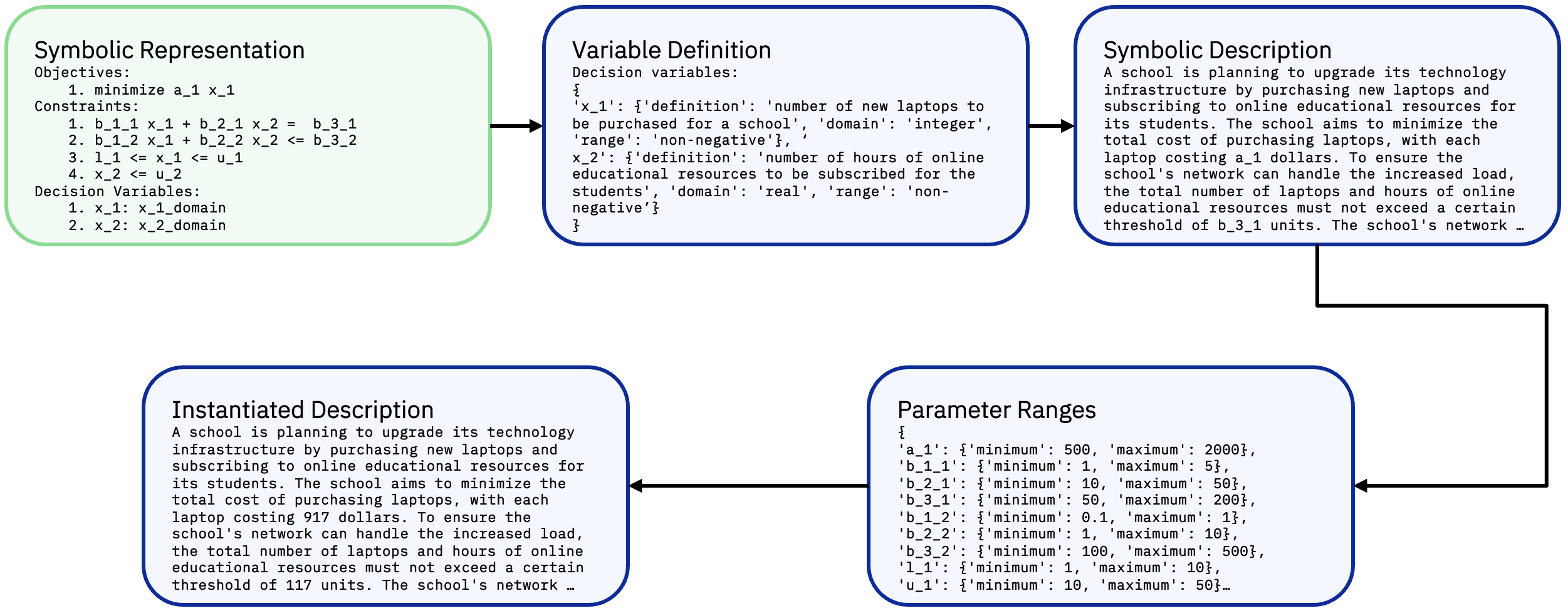}
\caption{An example illustrating the key steps from symbolic representation to problem description.}
\label{fig:sdg_example}
\end{figure}

Once the teacher model has successfully defined structured descriptions for all decision variables, we next prompt the model to generate sentences describing those variables, and to define value ranges for parameters associated with that variable. At this stage, we expect the teacher model to prepare a sentence describing the decision variables, but without instantiating any parameter values; for example,  
\begin{lstlisting}[numbers=none, xleftmargin=0pt, framexleftmargin=0pt, numbersep=0pt, breakindent=0pt]
A research institution needs to allocate resources to two key areas: purchasing microscopes and reagents, with a minimum of \\parameter{l_1} microscopes required.
\end{lstlisting}
The main motivation to do so is to condition parameter sampling on the context of the optimization problem, thereby avoiding potentially incoherent relationships -- for example, a negative budget. The SDG pipeline then iterates over the objective and constraints of the optimization problem in a similar manner. Once all components have been described, we prompt the teacher model to synthesize them into a single, uninstantiated problem statement, as shown by an example in Fig.~\ref{fig:sdg_example}. 
Moreover, if the implementation generated by the teacher model cannot be executed, but the teacher is able to debug that implementation successfully, we store that debugging step as a training sample. In the following, we empirically show that our approach improves performance of open-source models, providing a principled path for building reliable NL2Opt agents for real-world applications.

Next, we sample coefficients within the ranges defined by the teacher model, instantiate the parametrized optimization template, and select sets of coefficients that render the problem feasible. For feasible instances, we save the expected optimal value computed by the optimization template, and programmatically instantiate the symbolic natural language description generated by the teacher model. Because our work focuses on ensuring the correctness of mathematical formulations and code generated from natural language descriptions, we keep only instances where an optimal solution exists. In other words, we discard infeasible instances (where the feasible set is empty) and unbounded instances (where the feasible set is non-empty, but the objective value can be made arbitrarily good). We then pass the natural language description of each problem through the agentic workflow described in the previous section to generate trajectories that connect the problem description to solver-ready code. If (any of) the solutions found by the teacher model match the expected ground truth, we collect reasoning traces for each step of the workflow 
and save those for fine-tuning.
Each valid trajectory consists of three instruction–output pairs: 
\begin{itemize}
\small
\item[-] 
$\texttt{pair}_{\texttt{DA}}$\texttt{= (\{problem\_description, decomposition\_prompt\}; \{reasoning\_step\_1, extracted\_components\})}, 
\item[-] 
$\texttt{pair}_{\texttt{FA}}$\texttt{= (\{problem\_description, extracted\_components,  formulation\_prompt\}; \{reasoning\_step\_2, math\_formulation\})}, 
\item[-] 
$\texttt{pair}_{\texttt{CA}}$\texttt{= (\{problem\_description, math\_formulation,  coding\_prompt\}; \{reasoning\_step\_3, code in \{Pyomo, Gurobipy, DOcplex, CVXPY, PySCIPOpt\}\})}.
\end{itemize}

\paragraph{Tabular data} To improve diversity, we also generate problems with data stored in a table. 
We store the coefficients in tabular format, and prompt the teacher model to generate text descriptions for the row and column labels. We do not prompt the teacher model to prepare the tabular data; rather, we implement a Python script to define a string representing that table based on a set of feasible parameters and the labels generated by the teacher model. This allows us to randomly shuffle the sequence of decision variables and constraints, 
avoiding the generation of 
data in a fixed format. 

\paragraph{Extension to MILPs with abstract semantics} To extend the pipeline to classical MILPs such as traveling salesman or bin-packing problems --- in which the decision variables used in the formal model (whether to travel from a specific location to another, for example) do not exhibit a one-to-one correspondence with semantic decisions (the most efficient travel route) --- we define semantic proxies as more interpretable descriptions of the high-level decisions of the problem (e.g., locations to visit, or items that need to be packed), and prompt the teacher model to define those instead. We also provide the teacher model with a general description of the optimization class and indicate typical application domains. We employ this approach for traveling salesman, bin packing, multidimensional knapsack, set cover, and shift scheduling problems. We use a simplified generation procedure for certain structured problems such as maximum flow, minimum cost flow, and transportation network tasks. Speficically, we fix the number of decision variables, and rely on templated descriptions and paraphrasing for problem descriptions. While this restricts structural diversity, it enables us to efficiently scale data generation for these more complex problem classes. 

\paragraph{Teacher models} To generate varied problem descriptions, we rely on two models: Llama3.3-70B-Instruct \cite{grattafiori2024llama3herdmodels}, and Llama 4 Maverick \cite{llama4}. Similarly, to generate varied training demonstrations, we rely on three teacher models: Phi-4 \cite{abdin2024phi4technicalreport}, DeepSeek-R1-Distill-Llama-70B \cite{deepseekr1_2025}, and OpenAI o3-mini \cite{o3_mini}. Phi-4 and DeepSeek-R1-Distill-Llama-70B were primarily used to generate demonstrations for easier problems (linear problems, knapsack and multidimensional knapsack, set cover), and o3-mini for more challenging instances (e.g., traveling salesman, shift scheduling, bin packing, minimum cost flow). 

\paragraph{Multi-task Supervised Fine-Tuning}
To optimize performance, we train a small open-source LLM  (Granite 3.2 \cite{mishra2024granitecodemodelsfamily}) 
using targeted multi-instruction data synthesized by the 
pipeline described in the previous section. The training set includes detailed reasoning traces and diverse cross-format demonstrations designed to enhance generalization across tasks and representations. During fine-tuning, the model is provided with pairs $\texttt{pair}_{\texttt{DA}}, \texttt{pair}_{\texttt{FA}}$, and $ \texttt{pair}_{\texttt{CA}}$ of input instructions and corresponding reference outputs, typically in the form of \((\m{x}^{(i)}, \m{y}^{(i)})\) for \(i = 1, \ldots, N\) training examples. The loss function is the negative log-likelihood, which measures how well the model predicts each token in the reference output sequence. For a dataset of \(N\) instances, the loss is computed as:
\[
\mathcal{L}(\theta) = - \frac{1}{N} \sum_{i=1}^{N} \sum_{t=1}^{T^{(i)}} \log p_\theta\left(y_t^{(i)} \mid \m{x}^{(i)}, \m{y}_{0:t-1}^{(i)}\right)
\]
where \(T^{(i)}\) is the length of the \(i\)-th target sequence, \(y_t^{(i)}\) denotes the \(t\)-th token, and \(p_\theta\) is the model probability.


\section{Experimental Evaluation}

\begin{table*}[t]
\centering
\begin{tabular}{lccccccc}
\toprule
\textbf{Methods} & \textbf{NL4Opt} & \textbf{EasyLP} & \textbf{NLP4LP} & \textbf{ReSocratic} & \textbf{ComplexOR} & \textbf{IndustryOR} & \textbf{ComplexLP} \\
\midrule
GPT-4 & 61.2\% & 70.3\% & 73.6\% & 48.4\% & 42.9\% & 38.1\% & 57.7\% \\
CoT & 62.2\% & 49.5\% & 74.7\% & 43.6\% & 39.2\% & 40.5\% & 42.3\% \\
Chain-of-Experts & 66.7\% & \textbf{94.4}\% & 87.4\% & 71.2\% & 57.1\% & 31.2\% & 50.6\% \\
CAFA & 68.1\% & 71.2\% & 50.0\% & 40.1\% & 46.4\% & 41.1\% & 44.5\% \\
ORLM-LLaMA-3 8B & 73.8\% & 90.4\% & 76.4\% & 61.8\% & 50.0\% & \textbf{42.9\%} & 59.5\% \\
\textbf{OptiTrust} & \textbf{91.6\%} & 92.3\% & \textbf{94.4\%} & \textbf{81.4\%} & \textbf{61.1\%} & \textbf{42.9\%} & \textbf{63.1\%} \\
\bottomrule
\end{tabular}
\caption{Solution accuracy comparison across seven benchmark datasets.}
\label{tab:overall}
\end{table*}

To evaluate the effectiveness of our proposed \texttt{OptiTrust} agent, we compare it against state-of-the-art baseline methods (four prompting-based methods, including GPT-4 with standard prompting, Chain-of-Thought (CoT), Chain-of-Experts, and CAFA, and one fine-tuning-based method, ORLM-LLaMA-3 8B) on seven publicly available datasets commonly used in the optimization modeling literature: NL4Opt \cite{ramamonjison_nl4opt_2023}, EasyLP and ComplexLP \cite{huang_mamo_2024}, IndustryOR \cite{tang_orlm_2024}, NLP4LP \cite{ahmaditeshnizi_optimus_2024}, ComplexOR \cite{xiao_chain--experts_2023}, and ReSocratic \cite{yang2025optibench}. These datasets span different application domains and cover multiple types of optimization problems; however, as discussed in the recent survey paper \cite{Xiao_survey_2025}, the original datasets are unreliable for rigorous evaluation due to substantial inaccuracies arising from logical errors, poorly defined parameters, and incorrect ground truth data. Due to the inconsistency of ground-truth labels, as well as limited code accessibility, we compare performance primarily against the results reported in \cite{Xiao_survey_2025}. 
Our primary evaluation metric is Solution Accuracy, defined as the proportion of solutions correctly identifying the optimal solution to the optimization problems. This metric is widely accepted and recommended in prior benchmarking studies \cite{Xiao_survey_2025, yang2025optibench, huang_orlm_2025}. We use the same evaluation methodology in terms of the selection of baseline methods, cleaned datasets \cite{llm4or_repo}, and performance metric as proposed in \cite{Xiao_survey_2025}, which aims to establish the latest leaderboard for optimization modeling methods. 

We fine-tuned the Granite 3.2 8B Instruct model \cite{granite3_2} to serve as the agent backbone, 
leveraging 15000 synthetic training samples for a full fine-tuning - details 
are documented in the supplementary material. During training and evaluation, we permitted up to six iterations of debugging, as well as one self-reflection round for both decomposition and formulation agents.

\subsection{Overall Performance Analysis}

Table \ref{tab:overall} summarizes the solution accuracy of OptiTrust compared to baseline methods across all benchmark datasets. The training-free methods utilize the cutting-edge commercial OpenAI model gpt-4o-2024-08-06, ensuring state-of-the-art performance from prompting-based models. 

As presented in Table \ref{tab:overall}, our OptiTrust agent consistently achieves superior performance across the majority of datasets, attaining the highest solution accuracy in 6 out of the 7 evaluated benchmarks. Notably, OptiTrust significantly outperforms the next-best algorithm on several datasets such as NL4Opt and ReSocratic by at least 14\%, demonstrating its effectiveness in handling both standard and more nuanced optimization modeling challenges. While Chain-of-Experts shows strong performance on EasyLP, OptiTrust maintains a competitive second position, underscoring its robustness.

Our analysis further reveals that optimization problems characterized by complex and lengthy descriptions—such as ComplexOR, IndustryOR, and ComplexLP—remain particularly challenging for all evaluated methods, with solution accuracies consistently below 65\%. This indicates that substantial scope exists for further research into enhancing LLM agents' capability to interpret and accurately model highly complex optimization scenarios.

Overall, the empirical results strongly validate the efficacy and robustness of our proposed OptiTrust framework. By incorporating systematic, structured reasoning steps, majority voting, and leveraging verified synthetic training data, OptiTrust demonstrates significant potential for improving the reliability and interpretability of automated optimization modeling agents, setting a strong foundation for future advances in this important domain.


\subsection{Individual Modeling Language and Majority Voting Analysis}

We further investigate the performance for each modeling language and the effectiveness of majority voting across different optimization modeling languages within our OptiTrust framework. Specifically, we compare the solution accuracy of the OptiTrust agent employing majority voting across five popular optimization modeling languages: Pyomo, Gurobipy, PySCIPOpt, DOcplex, and CVXPY. Results using our OptiTrust agent with the base Granite model, without fine-tuning, are summarized in Table \ref{tab:baseGranite}, while results after fine-tuning Granite are presented in Table \ref{tab:FineTunedGranite}.

\begin{table*}[t]
\centering
\begin{tabular}{lcccccccc}
\toprule
\textbf{Methods} & \textbf{NL4Opt} & \textbf{EasyLP} & \textbf{NLP4LP} & \textbf{ReSocratic} & \textbf{ComplexOR} & \textbf{IndustryOR} & \textbf{ComplexLP} \\
\midrule
Pyomo      & 20.6\%  & 29.7\%  & 15.7\%  & 13.4\%  & 22.2\%   & 11.9\%  & 22.5\%  \\
Gurobipy   & \textbf{84.6\%}  & 79.4\%  & \textbf{88.2\%}  & \textbf{73.2\%}  & \textbf{38.9\%}   & \textbf{26.2\%}  & \textbf{37.8\%}  \\
PySCIPOpt  & 83.6\%  & \textbf{87.3\%}  & 84.3\%  & 68.7\%  & 22.2\%   & 23.8\%  & 35.1\%  \\
DOcplex    & 66.4\%  & 64.8\%  & 73.0\%  & 52.9\%  & 27.8\%   & 19.0\%  & 27.9\%  \\
CVXPY      & 43.9\%  & 57.8\%  & 36.5\%  & 41.4\%  & 0.0\%    & 14.3\%  & 29.7\%  \\
\midrule
OptiTrust  & 84.6\%  & 89.2\%  & 88.2\%  & 73.4\%  & 38.9\%   & 26.2\%  & 41.4\%  \\
\bottomrule
\end{tabular}
\caption{Solution accuracy comparison across modeling languages with non-fine-tuned Granite model.}
\label{tab:baseGranite}
\end{table*}

\begin{table*}
\centering
\begin{tabular}{lcccccccc}
\toprule
\textbf{Methods} & \textbf{NL4Opt} & \textbf{EasyLP} & \textbf{NLP4LP} & \textbf{ReSocratic} & \textbf{ComplexOR} & \textbf{IndustryOR} & \textbf{ComplexLP} \\
\midrule
Pyomo      & \textbf{90.7}\%  & 90.5\%  & 92.7\%  & 73.4\%  & 50.0\%   & \textbf{40.5}\%  & 51.4\%  \\
Gurobipy   & 87.9\%  & \textbf{91.6}\%  & \textbf{93.8}\%  & \textbf{79.9}\%  & \textbf{50.0}\%   & 35.7\%  & 56.8\%  \\
PySCIPOpt  & 87.9\%  & 91.4\%  & 91.6\%  & 76.9\%  & \textbf{50.0}\%   & 38.1\%  & 48.6\%  \\
DOcplex    & 89.3\%  & 91.0\%  & 92.7\%  & 75.7\%  & 44.4\%   & 28.6\%  & \textbf{57.7}\%  \\
CVXPY      & 89.3\%  & 89.0\%  & 92.7\%  & 73.9\%  & 27.8\%   & 28.6\%  & 45.9\%  \\
\midrule
OptiTrust  & 91.6\%  & 92.3\%  & 94.4\%  & 81.4\%  & 61.1\%   & 42.9\%  & 63.1\%  \\
\bottomrule
\end{tabular}
\caption{Solution accuracy comparison across modeling languages with fine-tuned Granite.}
\label{tab:FineTunedGranite}
\end{table*}

As demonstrated by these results, majority voting consistently enhances performance across all datasets. This aggregation mechanism effectively captures correct solutions by leveraging the diversity of solver implementations, thereby mitigating the limitations of any individual solver. Notably, substantial performance improvements are observed following the fine-tuning of the Granite model. Each modeling language demonstrates significant accuracy improvements after fine-tuning, underscoring the effectiveness of our synthetic data generation approach. For several datasets, including EasyLP and NLP4LP, the modeling languages achieve similarly strong performance levels, yielding comparably high accuracies after training. 

Moreover, it is noteworthy that more challenging datasets—such as ComplexOR, IndustryOR, and ComplexLP—exhibit greater relative improvements after fine-tuning. For instance, the accuracy for the ComplexOR dataset improved markedly from 38.9\% to 61.1\%, while ComplexLP increased from 41.4\% to 63.1\%, underscoring the enhanced capability of the fine-tuned model to handle complex optimization scenarios.

These findings illustrate the advantage of majority voting in enhancing robustness and solution accuracy, especially when combined with fine-tuning, highlighting the importance of integrating multiple modeling languages and rigorous training methodologies for complex optimization tasks.

\subsection{Addressing Data Quality Issues}
\label{sec:dataquality}

Existing optimization modeling datasets are significantly constrained in terms of scale, quality, and structural consistency. A key insight made by \cite{Xiao_survey_2025} highlights that the original 7 benchmark datasets are unreliable for rigorous evaluation due to substantial inaccuracies, primarily manifesting as incorrect optimal ground-truth values and logical inconsistencies within problem descriptions. Apart from the EasyLP dataset, the reported error rates in these benchmarks exceed 16\%, with the IndustryOR dataset exhibiting errors as high as 54\%. Such severe inaccuracies not only undermine the credibility of comparative studies but also emphasize the necessity for precise, verifiable, and robust datasets within the optimization modeling community.

To address this critical issue, \cite{Xiao_survey_2025} provided cleaned versions of these datasets by removing problematic data instances, resulting in a notable reduction in dataset size. Table \ref{tab:benchmark-quality} outlines the impact of this cleaning procedure, comparing the original number of problem instances (``Original Size") with the number remaining after cleaning (``Cleaned Size"). For datasets such as IndustryOR, ComplexLP, and ComplexOR, nearly half of the original data instances were excluded due to inaccuracies.

Building upon this previous effort, in this work, we further enhance the dataset quality beyond the initial cleaning process. We utilize our OptiTrust agent to systematically identify additional instances with incorrect ground truth optimal values. These inaccuracies were then corrected by using values from our agent, and the resulting revisions underwent rigorous validation by domain experts. As indicated in the Error Rate column of Table \ref{tab:benchmark-quality}, even after the previous cleaning efforts, some datasets such as NL4Opt and ComplexOR retained a non-trivial proportion of errors, highlighting the importance and effectiveness of our further refinements.

\begin{table}[t]
\small
\centering
\begin{tabular}{|l|r|r|r|}
\hline
\textbf{Dataset} & \textbf{Original Size} & \textbf{Cleaned Size} & \textbf{Error Rate} \\
\hline
NL4Opt     & 289 & 214 &  7.0\% \\
IndustryOR & 100 & 42 &  2.4\% \\
ComplexLP  & 211 & 111 &  2.7\% \\
ReSocratic & 605 & 405 &  0.7\% \\
ComplexOR  & 37  & 18 &  11.1\% \\
\hline
\end{tabular}
\caption{Detected and corrected error rates of optimization modeling benchmarks.}
\label{tab:benchmark-quality}
\end{table}

\section{Conclusion}

We have introduced \texttt{OptiTrust}, a modular LLM agent designed to translate natural language descriptions into solver-ready code by leveraging a principled synthetic data generation pipeline. By starting from a structured symbolic representation and systematically producing aligned natural language description, mathematical formulation, and code  with a verified optimal solution, our framework addresses the challenges of data scarcity, lack of verifiability, and limited generalization in optimization modeling with LLMs. Our approach enables fully programmatic construction of diverse and scalable NL2Opt datasets, supporting robust supervised fine-tuning across multiple modeling languages. The incorporation of stepwise demonstrations, multi-language inference, and majority-vote cross-validation leads to enhanced performance, as demonstrated by OptiTrust’s state-of-the-art results on a range of public benchmarks. Furthermore, our agent has proven valuable in identifying and correcting errors in existing datasets, thereby improving the reliability of evaluation standards within the community.

\appendix



\bigskip

\bibliography{nl2opt}

\newpage

\section{Appendix}
\subsection{A. \quad Data Generation}
We used the list of 18 problem domains suggested in \cite{ahmaditeshnizi_optimus_2024} as seed domains. The domains include: manufacturing and production, supply chain management, food and beverage, transportation and logistics, healthcare and medical, retail and e-commerce, environmental and sustainability, agriculture and forestry, science and research, energy and power systems, finance and banking, sports and entertainment, government and public sector, education, human resources, telecommunications, marketing and media, and aerospace and defense.

\subsection{B. \quad Cleaned Evaluation Datasets}

Our experimental evaluation utilizes a diverse set of benchmark datasets, each curated to assess the modeling, reasoning, and solver capabilities of large language models for mathematical optimization. The datasets include both academic and industrial benchmarks, spanning various optimization problem types, real-world scenarios, and difficulty levels. \cite{jiang2025llmopt} is one of the first papers to highlight the erroneousness of benchmark datasets, but it did not quantify the severity of the issue. Later, \cite{Xiao_survey_2025} systematically analyzes and reports that the majority of these datasets have error rates exceeding $20\%$, and provides a cleaned version. We further used our agent to improve the cleaned version. Below is a brief description of each original dataset and the final cleaned datasets.:

\begin{itemize}
    \item \textbf{NL4Opt:} NL4Opt originates from the NL4Opt Competition \cite{ramamonjison_nl4opt_2023}, designed to evaluate automated methods for translating optimization problems stated in natural language into code that can be processed by mathematical solvers. The original dataset primarily contains linear programs (LP) and mixed-integer linear programs (MILP) and features 289 instances. After filtering out low-quality examples by  \cite{Xiao_survey_2025}, it contains 214 instances. We further detected 15 incorrect ground truth values and fixed them for the processed dataset.

    \item \textbf{MAMO:} The MAMO dataset consists of two categories: \emph{EasyLP} and \emph{ComplexLP} \cite{huang_mamo_2024} with some nonlinear problems.
    \begin{itemize}
        \item \textbf{EasyLP:} Originally contains 652 instances covering a wide range of LP and MILP problems. After manually checking for correctness, the cleaned version remains 545 instances, our agent does not detect any inconsistencies. 
        \item \textbf{ComplexLP:} Provides 211 instances of higher difficulty, focusing on more complex problem statements. It consists of 2 nonlinear programs, and 65 combinatorial optimization problems. The cleaned dataset includes 111 instances, where our agent corrected 3 instances. 
    \end{itemize}
    
    \item \textbf{NLP4LP:} NLP4LP is sourced from the OptiMUS benchmark \cite{ahmaditeshnizi_optimus_2024}, originally comprising 344 linear and integer programming problems characterized by some lengthy descriptions and multi-dimensional parameters. After data cleaning, 178 consistent instances remain, with our agent detecting no inconsistencies.

    \item \textbf{ReSocratic:} It is the OptiBench comprehensive benchmark \cite{yang2025optibench} that encompasses both linear and nonlinear programming problems, including those with integer and mixed-integer variables. It consists of 605 original problems, covering a wide range of optimization contexts and tabular data. After cleaning, the dataset contains 405 instances, with our agent correcting three of them. 
    
    \item \textbf{ComplexOR:} ComplexOR, introduced in the Chain-of-Experts paper \cite{xiao_chain--experts_2023}, initially comprises 37 challenging optimization problems. The dataset features complex operations research scenarios—including three combinatorial optimization tasks—designed to assess the reasoning and problem-solving abilities of LLMs. After two corrections by our agent, the final number of instances is 18. 
    
    \item \textbf{IndustryOR:} IndustryOR is an industrial benchmark \cite{tang_orlm_2024} tailored for optimization modeling, introduced to evaluate the real-world applicability of LLMs. The dataset covers data from 13 different industries, 5 question types, and 3 levels of difficulty, reflecting a wide range of practical use cases. We corrected 1 instance, and the final number of cleaned instances is 42.
\end{itemize}

The final cleaned datasets are included in the zipped code files.


\subsection{C. \quad Implementation Details}

To fine-tune the models, we use NVIDIA H100 GPUs, with an effective batch size of 8 (four GPUs, each with a per device train batch size of 1 and 2 gradient accumulation steps). We set the learning rate to $1 \times 10^{-5}$, and use supervised fine-tuning recipes from the FMS HF Tuning repository \cite{fms_repo}, which relies on Hugging Face SFTTrainer \cite{vonwerra2022trl} and PyTorch FSDP \cite{zhao2023pytorchfsdpexperiencesscaling}. Moreover, we employ a two-stage supervised fine-tuning approach. In the first stage, we fine-tune the model for two epochs on a set consisting of 10000 linear problems (4000 with tabular data), 2000 knapsack problems (1000 with tabular data) and 2000 multidimensional knapsack problems (1000 with tabular data). To mitigate overfitting to synthetic data and promote generalization, we further incorporate 2000  training examples from GSM8K \cite{cobbe2021trainingverifierssolvemath}; 5000 instruction following (allenai/tulu-3-sft-personas-instruction-following) and 5000 code (allenai/tulu-3-sft-personas-code) samples from the T{\"u}lu 3 SFT dataset \cite{lambert2025tulu3pushingfrontiers}; as well as 9000 chain-of-thought samples from the Numina-CoT and Numina-TIR datasets \cite{numina_math_datasets}. In the second stage, we fine-tune the model for one additional targeted epoch using 1200 complex optimization problems, including traveling salesman, set cover, bin packing, shift scheduling, transportation, maximum flow, and minimum cost flow tasks.

We fine-tuned the Granite 3.2 8B Instruct model \cite{granite3_2} to serve as the agent backbone of our OptiTrust agent, 
leveraging 15000 synthetic training samples for a full fine-tuning. During training and evaluation, we permitted up to six iterations of debugging, as well as one self-reflection round for both the decomposition and formulation agents. During inference, we set the random seed to zero and the temperature to 0.7. We employed one NVIDIA H100 GPU via vLLM \cite{kwon2023efficient} for LLM inference and serving. We used the CPLEX solver for the modeling languages Pyomo, DOcplex, and CVXPY; the Gurobi solver for Gurobipy; and the SCIP solver for PySCIPOpt.

The accuracy is calculated based on our corrected datasets, presented in the Experimental Evaluation section, which are included in the zipped file. The condition to verify the correctness of the optimal value is $|f_{\text{OptiTrust}} - f_{\text{label}}| \le \epsilon$, where $\epsilon = 10^{-4}$. Here, $f_{\text{OptiTrust}}$ is the value produced by our agent, and $f_{\text{label}}$ is the ground truth. We observed that for some instances, the ground truth values are rounded to one decimal place; in such cases, we used $\epsilon = 10^{-1}$ instead. 

\subsection{D. \quad Additional Numerical Results}

\begin{table*}
\centering
\begin{tabular}{lccccccc}
\toprule
\textbf{Methods} & \textbf{NL4Opt} & \textbf{EasyLP} & \textbf{NLP4LP} & \textbf{ReSocratic} & \textbf{ComplexOR} & \textbf{IndustryOR} & \textbf{ComplexLP} \\
\midrule
Pyomo      & \textbf{37.4}\%   & 55.2\%   & 66.3\%   & \textbf{30.5}\%   & 0.0\%   & \textbf{28.6}\%   & 12.6\%  \\
Gurobipy   & 31.3\%   & \textbf{63.9}\%   & \textbf{67.4}\%   & 30.3\%   & 0.0\%   & 21.4\%   & \textbf{14.4}\%  \\
PySCIPOpt  & 34.6\%   & 62.4\%   & 60.7\%   & 30.3\%   & 0.0\%   & 23.8\%   & 5.4\%  \\
DOcplex    & 33.2\%   & 57.2\%   & 56.2\%   & 25.8\%   & \textbf{5.6}\%   & 26.2\%   & 13.5\%  \\
CVXPY      & 35.5\%   & 54.7\%   & 64.6\%   & 29.8\%   & 0.0\%   & 23.8\%   & 9.9\%  \\
\midrule
OptiTrust  & 49.1\%   & 67.9\%   & 75.8\%   & 45.4\%   & 5.6\%   & 31.0\%   & 15.3\%  \\
\bottomrule
\end{tabular}
\caption{Solution accuracy comparison across modeling languages for Qwen 1.5 14B  without fine-tuning. For each dataset, the highest score among Pyomo, Gurobipy, PySCIPOpt, DOcplex, and CVXPY is shown in bold.}
\label{tab:SolutionAccuracyQwen}
\end{table*}

\begin{table*}
\centering
\begin{tabular}{lccccccc}
\toprule
\textbf{Methods} & \textbf{NL4Opt} & \textbf{EasyLP} & \textbf{NLP4LP} & \textbf{ReSocratic} & \textbf{ComplexOR} & \textbf{IndustryOR} & \textbf{ComplexLP} \\
\midrule
Pyomo      & \textbf{85.0}\%   & \textbf{85.0}\%   & \textbf{87.6}\%   & \textbf{71.2}\%   & 27.8\%    & 35.7\%   & 60.4\%  \\
Gurobipy   & 75.2\%   & 76.0\%   & 79.2\%   & 67.0\%    & 44.4\%    & 31.0\%   & 63.1\%  \\
PySCIPOpt  & 78.0\%   & 79.8\%   & 82.6\%   & 66.5\%    & \textbf{44.4}\%    & \textbf{35.7}\%   & \textbf{63.1}\%  \\
DOcplex    & 80.4\%   & 81.7\%   & 84.3\%   & 69.5\%    & \textbf{44.4}\%    & 31.0\%   & 61.3\%  \\
CVXPY      & 84.1\%   & 79.6\%   & 86.0\%   & 67.0\%    & 27.8\%    & 28.6\%   & 45.9\%  \\
\midrule
OptiTrust  & 88.3\%   & 91.2\%   & 91.0\%   & 77.9\%    & 50.0\%    & 40.5\%   & 77.5\%  \\
\bottomrule
\end{tabular}
\caption{Solution accuracy comparison across modeling languages for Qwen 1.5 14B  with fine-tuning. For each dataset, the highest score among Pyomo, Gurobipy, PySCIPOpt, DOcplex, and CVXPY is shown in bold.}
\label{tab:SolutionAccuracyQwenSFT}
\end{table*}

\begin{table*}[t]
\centering
\begin{tabular}{lcccccccc}
\toprule
\textbf{Methods} & \textbf{NL4Opt} & \textbf{EasyLP} & \textbf{NLP4LP} & \textbf{ReSocratic} & \textbf{ComplexOR} & \textbf{IndustryOR} & \textbf{ComplexLP} \\
\midrule
\textit{\textbf{Execution Rate}} \\
\midrule
Qwen non-fine-tuned  & 98.1\%  &	\textbf{99.6\%}  &	\textbf{100.0\%}  &	98.8\%  &	72.2\%  &	\textbf{100\%}  &	97.3\% \\
Qwen fine-tuned  & 99.5\%  &	98.0\%  &	\textbf{100.0\%}  &	97.5\%  &	77.8\%  &	95.2\%  &	98.2\% \\
Granite non-fine-tuned  & \textbf{100.0\%}  &	99.1\%  &	99.4\%  &	96.8\%  &	50.0\%  &	81.0\%  &	85.6\%  \\
Granite fine-tuned  &\textbf{100.0\%}  &	\textbf{99.6\%}  &	\textbf{100.0\%}  &	\textbf{99.8\%}  &	\textbf{88.9\%}  &	90.5\%  &	\textbf{99.1\%} \\ 
\midrule
\textit{\textbf{Accuracy Rate}} \\
\midrule
Qwen non-fine-tuned  & 49.1\%   & 67.9\%   & 75.8\%   & 45.4\%   & 5.6\%   & 31.0\%   & 15.3\%  \\
Qwen fine-tuned  & 88.3\%   & 91.2\%   & 91.0\%   & 77.9\%    & 50.0\%    & 40.5\%   & \textbf{77.5\%}  \\
Granite non-fine-tuned   & 84.6\%  & 89.2\%  & 88.2\%  & 73.4\%  & 38.9\%   & 26.2\%  & 41.4\%  \\
Granite fine-tuned  & \textbf{91.6\%}  & \textbf{92.3\%}  & \textbf{94.4\%}  & \textbf{81.4\%}  & \textbf{61.1\%}   & 42.9\%  & 63.1\%  \\
\bottomrule
\end{tabular}
\caption{Execution rate and accuracy rate comparison for non-fine-tuned and fine-tuned Qwen and Granite models.}
\label{tab:ExecutionRate}
\end{table*}

In this section, experiments for base and fine-tuned Qwen 1.5 14B models \cite{qwen15_14B} for OptiTrust are summarized in Tables \ref{tab:SolutionAccuracyQwen} and \ref{tab:SolutionAccuracyQwenSFT}, respectively. In general, the base Qwen model achieves worse accuracy rates than the non-finetuned Granite model. We conjecture that the difference in performance is due to Granite 3.2 8B being a more recent model, on the one hand, and being exposed to supervised instruction data that enables it to better utilize the provided in-context learning demonstrations, on the other. It is also worth noting that the discrepancy in accuracy across modeling languages is smaller for the base Qwen 1.5 14B model than for the non-fine-tuned Granite model.

We also report both the accuracy rate and the execution rate for the non-fine-tuned and fine-tuned Qwen 1.5 14B and Granite 3.2 8B Instruct models on OptiTrust. The execution rate is defined as the proportion of problem instances whose code runs without errors and generates output results. As shown in Table~\ref{tab:ExecutionRate}, our multi-language fine-tuning data for OptiTrust results in higher accuracy rates, especially for complex problems in the ComplexOR, IndustryOR, and ComplexLP datasets. This highlights the effectiveness of the synthetic data generation method. Overall, we also observe improvements in execution rate after fine-tuning, except for the Qwen model on three datasets: EasyLP, ReSocratic, and IndustryOR.

\onecolumn
\newpage
\subsection{E. \quad Prompt Templates}
 In this section, we present the key prompt templates. Owing to space limitations, we provide only the structures of the prompts, omitting some details. Additional prompts are available in the zipped code.

\subsection{E.1. \quad OptiTrust Agent Prompt}

This subsection describes an end-to-end prompting workflow to decompose, formulate, and implement optimization problem.

\begin{prompttemplateWhite}{Decomposition Agent Prompt}
You are an expert in mathematical optimization. Your task is to identify and prepare natural language descriptions of components of an optimization problem.

Upon receiving a problem description, you should:

1. Carefully analyze and comprehend the problem.
2. Summarize the decision variables related to the problem. Indicate whether each of the decision variables is required to be integer, real or binary based on the context of the problem.
3. Summarize and define the objective of the problem. Indicate any parameters or numerical values needed to define the objective. 
4. Identify and list all constraints, including any implicit ones like non-negativity. List and summarize the constraints using natural language. Indicate any parameters or numerical values needed to define each of the constraints
5. Verify if any numerical values or parameters defined in the problem description are missing from the objective or constraints you identified, and update the list of components you prepared, if necessary.

Note that
- If adding any mathematical expressions, try to mathematically represent constraints and objectives as close to their natural language description as possible; you do not need to simplify any constraints or objectives.
- The final list of components should be enclosed between the "```" lines.

Here is a description of the problem we need you to find the components for:
-----
{description}
-----

Now, follow the steps outlined above. Explain your reasoning and remember to enclose the final list of components between the "```" lines.
\end{prompttemplateWhite}


\begin{prompttemplateWhite}{Decomposition Verifier Prompt}
You are an expert in mathematical optimization. Your task is to review previously identified components of an optimization problem.

Upon receiving the description of an optimization problem and a list of previously identified components of the optimization problem, you should:

1. Carefully analyze and comprehend the problem.
2. Verify if the decision variables, objectives and constraints listed in the previously prepared list of components have been identified correctly. 
3. Verify if any decision variables, objectives or constraints in the description of the optimization problem are missing from the previously prepared list of components and update the list, if necessary. 
4. Verify if any numerical values or parameters defined in the problem description are missing from the components you identified, and update the list of components you prepared to include those, if necessary.
5. Prepare a final, revised list with the components of the optimization problem (including objectives, constraints and decision variables) in natural language. Make sure to avoid repeating components.

Note that
- You should include any implicit constraints such as non-negativity
- If adding any mathematical expressions, try to mathematically represent constraints and objectives as close to their natural language description as possible; you do not need to simplify any constraints or objectives.
- You should indicate whether each of the decision variables is required to be integer, real or binary based on the context of the problem.
- The final list of components should be enclosed between the "```" lines. 

Here is a description of the problem we need you to find the components for:
-----
{description}
-----

And here is the list of previously identified components:
-----
{previous_components}
-----

Now, follow the steps outlined above. Explain your reasoning and remember to enclose the final list of components between the "```" lines.
\end{prompttemplateWhite}


\begin{prompttemplateWhite}{Formulation Agent Prompt}
You are an expert in mathematical optimization, and your task is to model an optimization problem.

Upon receiving the description of an optimization problem, you should:

1. Carefully analyze and comprehend the problem. 
2. Carefully review the decision variables previously identified. Define symbols representing the decision variables and indicate whether each of the decision variables is required to be integer, real or binary based on the context of the problem. 
3. Indicate whether any decision variables are required to be non-negative based on the context of the problem.
4. Carefully review the previously identified objectives, and prepare a mathematical formulation representing the objective. If the optimization problem has multiple objectives, convert a multi-objective optimization problem into a single-objective optimization problem using linear scalarization with the weights of the objectives.
5. Carefully review the previously identified constraints, and prepare a mathematical formulation representing each of the constraints.
6. Prepare a mathematical formulation of the problem using LaTeX.
7. Verify if any numerical values or parameters defined in the problem description are missing from the formulation, and update the mathematical formulation to include them, if necessary.

Note that
- Try to mathematically represent constraints and objectives as close to their natural language description as possible.
- You do not need to simplify any constraints or objectives.
- Your formulation should be in LaTeX mathematical format. 
- The final mathematical formulation should be enclosed between the "```" lines.

Here is a description of the problem we need you to model:
-----
{description}
-----

The following components have been previously identified:
-----
{components}
-----

Now, solve the problem step by step. Explain your reasoning and remember to enclose the final list of components between the "```" lines.
\end{prompttemplateWhite}

\begin{prompttemplateWhite}{Formulation Verifier Prompt}
You're an expert in mathematical optimization. You need to revise the mathematical formulation of an optimization problem prepared by a student.

Here is a description of the problem we need you to model:
-----
{description}
-----

The following components have been previously identified:
-----
{components}
-----

And here is the mathematical formulation we need you to verify:
-----
{previous_formulation}
-----

Solve the problem step by step. Explain your reasoning and remember to enclose the final list of components between the "```" lines.

\end{prompttemplateWhite}

\begin{prompttemplateWhite}{Programmer Prompt}
You are an expert in mathematical optimization, and your objective is to create a Python script to solve an optimization problem using {solver}. When solving an optimization problem, you should follow a structured approach:
1. Carefully analyze and comprehend the problem description.
2. Carefully analyze and comprehend the provided decomposition of the problem into a detailed list of decision variables, objective(s) and constraints.
3. Carefully analyze and comprehend the previously prepared mathematical formulation of the optimization problem.
4. Prepare a well-documented Python script to solve the optimization problem using {solver}. Anchor your implementation on the context, the detailed decomposition, and the formulation of the optimization problem. Pay special attention to the domain of each decision variable, implicit constraints such as non-negativity, and that all relevant parameters are included in the script you generate.
Note that
- You should clearly explain your reasoning and the steps you take to solve the problem.
- You should enclose the final code between "```" lines, as in the provided examples.
- You should print the optimal value of the optimization problem using 'Optimal value: ', as in the provided examples.
Let's think step by step and clearly describe our reasoning.


Here is the problem description:
-----
{description}
-----

The following components have been extracted from the problem description:
-----
{components}
-----

And the following mathematical formulation to represent the optimization problem has been prepared:
-----
{formulation}
-----

Now, follow the steps outlined above. Explain your reasoning and remember to enclose the generated code between "```" lines.  
\end{prompttemplateWhite}

\begin{prompttemplateWhite}{Code Debugging Prompt}
Your task is to debug Python {solver} code used to solve the following optimization problem:
-----
{description}
-----
This is the code snippet for which an error occured
-----
{code_w_error}
-----
and here is the error message:
-----
{error_message}
-----

First reason about the source of the error, and decide whether it is a modeling issue, or a code bug. Then, generate a Python {solver} script accordingly to fix any errors, and enclose it between "```" lines.
\end{prompttemplateWhite}

\begin{prompttemplateWhite}{Infeasibility Debugging Prompt}
Your task is to investigate {solver} code used to solve the following optimization problem:
-----
{description}
-----
This is the code snippet initially used to solve the problem:
-----
{code_w_error}
-----
That optimization model, however, is infeasible. Your task now is to investigate the source of that issue, and decide whether infeasibility is due to a code bug or a modeling issue. Then, generate a Python {solver} script accordingly to fix any errors or modeling issues, and enclose it between "```" lines.
Now, here is a demonstration showing how to solve this task:
-----
{code_examples}
-----
Now, reason about the source of the error, generate a Python {solver} script accordingly to fix any issues, and enclose it between "```" lines.
\end{prompttemplateWhite}

\subsection{E.2. \quad Synthetic Data Generation Prompt}

It generates a description of an optimization problem based on symbolic representation
        of an optimization problem. The SDG workflow consists of (i) sampling an
        application domain, (ii) prompting teacher model to create a new seed (with
        the underlying objective of improving natural language diversity), (iii) prompting teacher model to define decision variables, (iv) prepare a new problem description, and (v) define ranges for each parameter in the optimization problem.

\begin{prompttemplateWhite}{A Natural Language Description Seed Prompt}
You are an expert in mathematical optimization, and your task is to create a new optimization problem. Upon receiving some sample problem descriptions, you should create a new optimization problem within the {industry} domain and with {no_variables} variables that follows a structure similar to the provided samples.

Note that
- The new optimization problem should be enclosed between the "```" lines.
- The problem you generate should be new, that is, do not repeat any of the provided examples.
- Avoid using symbols in the problem description you generated, try to use natural language instead. For example, instead of saying "A manufacturing company produces two products, A and B", try to say something like "A manufacturing company produces two products, soap bars and shampoo bottles".
- Make sure to use the {industry} domain in your new optimization problem.

Here's a demonstration showing how to complete your task
{q_n_a_sample}
and here are some sample problem descriptions to base the new problem description on
{sample_problems}
\end{prompttemplateWhite}

\begin{prompttemplateWhite}{Variable Definition Prompt}
You are an expert in mathematical optimization, and your task is to define a list of decision variables for an optimization problem. Upon receiving a pre-defined list of decision variables and a pre-initialized description of an optimization problem, you should

1. Carefully analyze and comprehend the provided formulation.
2. Clearly define the decision variables present in the pre-initialized description. Be specific.
3. Define the domain of those variables, that is, indicate whether those variables are continous, binary or integer based on the context of the problem. Indicate whether any decision variable should be non-negative or not.

Note that your final response should be between the ``` lines, as shown in the provided demonstration.

Here's a demonstration showing how to complete your task
-----
{qna_examples}
-----

Now, here are the decision variables we need you to prepare a detailed list for:
-----
{formulation}
-----
and here is the pre-initialized description:
-----
{description}
-----
Now, follow the steps above, generate a list indicating the decision variables in the optimization problem.
\end{prompttemplateWhite}

\begin{prompttemplateWhite}{Variable Definition Debugging Prompt}
You are a mathematical optimization expert. Your task is to review a list of decision variables prepared for a new optimization problem. Upon receiving a detailed mathematical formulation of an optimization problem and the variables missing from the definition, you should:

1. Carefully analyze and comprehend the provided formulation.
2. Carefully review the previously prepared description of decision variables for the provided formulation.
3. Carefully revise the previously prepared list describing the decision variables, and refine it to include the missing variables.

Note that
- The list of variables should be enclosed between the "```" lines.

Here's a demonstration showing how to complete your task
-----
{qna_examples}
-----

Now, here is the formulation of the optimization problem we need you to prepare a list of decision variables for:
-----
{formulation}
-----
here is the problem description:
-----
{description}
-----
and here is the list of missing variables:
-----
{missing_components}
-----
\end{prompttemplateWhite}

\begin{prompttemplateWhite}{Objective Definition Prompt}
You are an expert in the {industry} domain working closely with a mathematical optimization professor to prepare realistic descriptions of optimization problems for a new textbook. Your task is to first think of a unique aspect of a business case within the {industry} domain, then indicate a real metric with which that goal can be measured (such as dollars, seconds, minutes, hours, miles, pounds, cubic meters, ml, etc), and then create a realistic natural-language description that reflects the objective of the new optimization problem generated by the optimization expert.

Here's a demonstration showing the expected format
-----
{q_n_a_sample}
-----
and here is the detailed formulation of the problem you should base the description you generate on
-----
{formulation}
-----

Note that
- The objective description should be enclosed between the "```" lines, and focus on a clear, measurable aspect of a business case within the {industry} domain.
- All parameters present in the formulation should be written in the form \\parameter in the corresponding natural language description, as in the provided examples.
- Do not include symbols from the provided formulation into the description you generate. That is, do not include symbols such as "x_1" in the description you generate.
- Do not initialize the parameter values indicated in the problem description.
- Don't be explicit about non-negative quantities or the type of decision variable, that is, do not include statements such as "The number of libraries must be a non-negative integer" in the description you generate.
- Carefully review the description you generated, and make sure it includes all the decision variables and parameters present in the optimization objective.
- Do not add any new variables or parameters not listed in the provided formulation.
- Don't simply write a literal description of the objective, try to create a realistic description of a business case within your {industry} expertise.
- For any negative parameters, try to indicate the non-negativity by using terms such as "penalty", "loss", "decreases", "reduces".
\end{prompttemplateWhite}

\begin{prompttemplateWhite}{Objective Definition Debugging Prompt}
You are an expert in the {industry} domain working closely with a mathematical optimization professor to prepare realistic descriptions of optimization problems for a new textbook. Your task is to revise the description of the objective of a new optimization problem to include a list of missing parameters.

Here's a demonstration showing the expected format
-----
{q_n_a_sample}
-----
here is the detailed formulation of the problem you should base the description you generate on
-----
{formulation}
-----
here is the description we need you to revise:
-----
{previous_description}
-----
and, finally, here is the list of missing parameters
-----
{missing_params}
-----

Note that
- The revised description should be enclosed between the "```" lines, and focus on a clear, measurable aspect of a business case within the {industry} domain.
- All parameters present in the formulation should be written in the form \\parameter in the corresponding natural language description, as in the provided examples.
- Do not include symbols from the provided formulation into the description you generate. That is, do not include symbols such as "x_1" in the description you generate.
\end{prompttemplateWhite}

\begin{prompttemplateWhite}{Constraint Definition Prompt}
You are an expert in the {industry} domain working closely with a mathematical optimization professor to prepare realistic descriptions of optimization problems for a new textbook. Your task is to first think of a unique aspect of a business case within the {industry} domain, then indicate a real, well-defined metric with which that goal can be measured (such as dollars, seconds, minutes, hours, miles, pounds, cubic meters, ml, etc), and then create a realistic natural-language description that reflects one of the constraints of the new optimization problem generated by the optimization expert.

Here's a demonstration showing the expected format
-----
{q_n_a_sample}
-----
and here is the detailed formulation of the problem you should base the description you generate on
-----
{formulation}
-----
The following topics have already been used, so try to focus on a new aspect:
-----
{previous_topics}
-----

Note that
- The constraint description should be enclosed between the "```" lines, and focus on a clear, measurable aspect of a business case within the {industry} domain.
- All parameters present in the formulation should be written in the form \\parameter in the corresponding natural language description, as in the provided examples.
- Do not include symbols from the provided formulation into the description you generate. That is, do not include symbols such as "x_1" in the description you generate.
- Do not initialize the parameter values indicated in the problem description.
- Don't be explicit about non-negative quantities or the type of decision variable, that is, do not include statements such as "The number of libraries must be a non-negative integer" in the description you generate.
- Carefully review the description you generated, and make sure it includes all the decision variables and parameters present in the constraint.
- Do not add any new variables or parameters not listed in the provided formulation.
- Don't simply write a literal description of the constraint, try to create a realistic description of a business case within your {industry} expertise.
- For any negative parameters, try to indicate the non-negativity by using terms such as "penalty", "loss", "decreases", "reduces".
\end{prompttemplateWhite}

\begin{prompttemplateWhite}{Constraint Definition Debugging Prompt}
You are an expert in the {industry} domain working closely with a mathematical optimization professor to prepare realistic descriptions of optimization problems for a new textbook. Your task is to revise the description of a constraint of a new optimization problem to include a list of missing parameters.

Here's a demonstration showing the expected format
-----
{q_n_a_sample}
-----
here is the detailed formulation of the problem you should base the description you generate on
-----
{formulation}
-----
here is the description we need you to revise:
-----
{previous_description}
-----
and, finally, here is the list of missing parameters
-----
{missing_params}
-----

Note that
- The revised description should be enclosed between the "```" lines, and focus on a clear, measurable aspect of a business case within the {industry} domain.
- All parameters present in the formulation should be written in the form \\parameter in the corresponding natural language description, as in the provided examples.
- Do not include symbols from the provided formulation into the description you generate. That is, do not include symbols such as "x_1" in the description you generate.
\end{prompttemplateWhite}

\begin{prompttemplateWhite}{Parameter Range Definition Prompt}
You are a mathematical optimization expert. Your task is to define a list of plausible value ranges for parameters in an optimization problem. Upon receiving a detailed mathematical formulation of an optimization problem, and its natural language description, you should:

1. Carefully analyze and comprehend the provided formulation.
2. Carefully analyze and comprehend the problem description within the {industry} domain.
3. Define suitable minimum and maximum values for all listed parameters within the {industry} domain.

Note that
- The range of values should be enclosed between the "```" lines. 
- Make sure to use suitable values within the {industry} domain.
- Make sure to enclose the range of values between the "```" lines.

Here is the formulation of the optimization problem 
-----
{formulation}
-----
and here is the natural language description of the problem:
-----
{description}
-----

Now, follow the steps outlined above. The range of values should be enclosed between the "```" lines. Here are some examples:
-----
{qna_examples}
-----

Solve the problem step by step.
\end{prompttemplateWhite}

\begin{prompttemplateWhite}{Parameter Range Debugging Prompt}
You are a mathematical optimization expert. Your task is to review a list of plausible value ranges for parameters in an optimization problem. Upon receiving a detailed mathematical formulation of an optimization problem, its natural language description, and a list of parameters for which minimum and maximum values were not defined, you should:

1. Carefully analyze and comprehend the provided formulation.
2. Carefully analyze and comprehend the problem description within the {industry} domain.
3. Carefully review the previously prepared range of plausible values for parameters included in the provided formulation.
4. Carefully revise the previously prepared range of plausible values for all parameters, and refine it to include the missing parameters.
5. Define suitable minimum and maximum values for all listed parameters within the {industry} domain.

Note that
- The range of values should be enclosed between the "```" lines. 
- Make sure to use suitable values within the {industry} domain.
- Make sure to enclose the range of values between the "```" lines.

Here is the formulation of the optimization problem 
-----
{formulation}
-----
here is the natural language description of the problem
-----
{description}
-----
here is the definition of parameter ranges we need you to revise:
-----
{previous_description}
-----
and here is the list of missing parameters:
-----
{missing_components}
-----

Now, follow the steps outlined above. The range of values should be enclosed between the "```" lines. Here are some examples:
-----
{qna_examples}
-----

Solve the problem step by step.
\end{prompttemplateWhite}

\begin{prompttemplateWhite}{Natural Language Description Prompt}
You are an editor working closely with a mathematical optimization professor on a new textbook, and your task is to create a natural-language description for a new problem for the textbook. Upon receiving a detailed list of decision variables, objectives, and constraints prepared by the optimization professor, you should create a well-written description that harmoniously integrates all the aspects of the problem previously prepared by the optimization expert.

Here's a demonstration showing the expected format
-----
{q_n_a_sample}
-----
and here is the detailed list of decision variables, objective, and constraints you should include in the new description
-----
{formulation}
-----

Note that
- The final description of the optimization problem should be enclosed between the "```" lines. 
- All parameters present in the formulation should be written in the form \\parameter in the corresponding natural language description, as in the provided examples. But do not add parenthesis to the parameters, that is, do not use \( \) in the description you generate.
- Do not initialize the parameter values indicated in the problem description.
- Don't be explicit about non-negative quantities or the type of decision variable, that is, do not include statements such as "The number of libraries must be a non-negative integer" in the description you generate.
- Carefully review the description you generated, and make sure it includes all the decision variables, constraints and objectives of the provided formulation.
- The description should not include any decision variables or parameters not listed in the provided formulation.
\end{prompttemplateWhite}

\begin{prompttemplateWhite}{Symbolic Debugging Prompt}
You are a mathematical optimization expert. Your task is to review a description prepared for a new optimization problem. Upon receiving a detailed mathematical formulation of an optimization problem and a list of parameters missing from the description, you should:

1. Carefully analyze and comprehend the provided formulation.
2. Carefully review the previously prepared description for the provided formulation.
3. Carefully revise the previously prepared description, and refine it to include the missing parameters.
4. Make sure that all parameters are written in the form \\parameter, as in the provided examples.

Note that
- The description of the optimization problem should be enclosed between the "```" lines. 
- The description you generate should be new, that is, do not repeat any of the provided examples.
- Avoid using symbols representing the decision variables in the problem description you generate, use natural language instead. For example, instead of saying "A manufacturing company produces two products, A and B", try to say something like "A manufacturing company produces two products, soap bars and shampoo" 
- Make sure to use the {industry} domain in the description of the optimization problem.
- Make sure to use meaningful measures for parameters included in the description of the optimization problem. That is, instead of using something like "at least b_1_1 units", try saying "at least b_1_1 gallons".
- Do not include symbols from the provided formulation into the description you generate. That is, do not include symbols such as "x_1" in the description you generate.
- Do not include mathematical expressions in your description. Use natural language instead. For example, do not include something like $x_3 - x_4 \ge 10$ in the description you generate.
- Do not initialize the parameter values indicated in the problem description.
- Don't be explicit about non-negative quantities, that is, do not include statements such as "The number of libraries must be a non-negative integer" in the description you generate.
- Don't be explicit about the type of decision variables, that is, do not include statements such as "The number of libraries must be an integer" in the description you generate.
- Only generate the description of the optimization problem, and nothing else.  
- Make sure to enclose the problem description between the "```" lines.

Here is the formulation of the optimization problem
-----
{formulation}
-----
here is the description we need you to revise:
-----
{previous_description}
-----
and here is the list of missing components:
-----
{missing_components}
-----

Now, follow the steps outlined above. The problem description should be enclosed between the "```" lines. Here are some examples:
-----
{qna_examples}
-----

Solve the problem step by step. Be concise.
\end{prompttemplateWhite}


\end{document}